\documentclass[letterpaper, 10 pt, conference]{ieeeconf}  %

\IEEEoverridecommandlockouts                              %

\usepackage{amsmath} %
\usepackage{amssymb}  %

\usepackage{graphicx}
\usepackage{cite}
\usepackage{booktabs}
\usepackage{xcolor} 
\definecolor{darkgreen}{rgb}{0.0, 0.5, 0.0}
\definecolor{darkred}{rgb}{0.55, 0.0, 0.0} %
\usepackage{array}    %
\usepackage{booktabs} 
\usepackage{adjustbox}
\usepackage{tabularx}
\usepackage{booktabs}
\usepackage{array}
\usepackage{threeparttable}

\title{\LARGE \bf
MistyPilot: An Agentic Fast–Slow Thinking LLM Framework for Misty Social Robots
}

\author{%
Xiao Wang$^{*}$, Lu Dong$^{*}$, Jingchen Sun, Ifeoma Nwogu, Srirangaraj Setlur, Venu Govindaraju%
\thanks{$^{*}$These authors contributed equally.}%
\thanks{All authors are with the State University of New York at Buffalo, Buffalo, NY 14260, USA.
Emails: \{xwang277, ludong, jsun39, inwogu, setlur, govind\}@buffalo.edu}%
}

\begin{document}

\maketitle
\thispagestyle{empty}
\pagestyle{empty}

\begin{abstract}

With the availability of open APIs in social robots, it has become easier to customize general-purpose tools to meet users’ needs. However, interpreting high-level user instructions, selecting and configuring appropriate tools, and executing them reliably remain challenging for users without programming experience.
To address these challenges, we introduce MistyPilot, an agentic LLM-driven framework for autonomous tool selection, orchestration, and parameter configuration. 
MistyPilot comprises two core components: a \textit{Physically Interactive Agent (PIA)} and a \textit{Socially Intelligent Agent (SIA)}. 
The \textit{PIA} enables robust sensor-triggered and tool-driven task execution, while the \textit{SIA} generates socially intelligent and emotionally aligned dialogue.
MistyPilot further integrates a fast–slow thinking paradigm to capture user preferences, reduce latency, and improve task efficiency.
To comprehensively evaluate MistyPilot, we contribute five benchmark datasets. Extensive experiments demonstrate the effectiveness of our framework in routing correctness, task completeness, fast–slow thinking retrieval efficiency, tool scalability, and emotion alignment. All code, datasets, and experimental videos will be made publicly available on the project webpage.

\end{abstract}

\section{INTRODUCTION}

Social robots, as emerging stars of the modern era, are increasingly involved in education, daily living, elder care, and emotional support, demonstrating advanced companionship and assistive capabilities \cite{wang2025social,jackson2009teachers,gonzalez2025social, romero2024using, sawik2023robots, zhao2025social}.
Misty, a compact, intelligent, and elegantly designed social robot, is increasingly adopted for both interactive and assistive roles \cite{wang2025automisty, ciuffreda2025design, eckert2025programming}. It offers atomic-level open APIs—including tactile sensing, articulated head and arm control, and camera and speech modules—that support the development of versatile functional tools such as sensor-triggered greetings, camera activation, content retrieval, conversational question answering, emotional support, and daily memos.
However, high-level user instructions are often abstract, diverse, and inherently uncertain, making it impractical to enumerate and hard-code all possible scenarios. For example, a user might request a story about 'The Three Little Pigs' and later wish to change the ending and hear it retold. A promising strategy to address such challenges is to pre-define a set of general, molecular-level tools that can be flexibly composed as needed. 
Much like possessing tools, leveraging them effectively presents its own challenges; Nevertheless, selecting and configuring appropriate tools remains a significant challenge for non-expert users.

As user demands continue to evolve, scalability is essential for social robots. However, most existing systems lack plug-and-play functionality, requiring extensive manual coding and configuration to add or integrate tools. This rigid design limits adaptability to new tasks and diverse user needs. Therefore, next-generation social robots should enable seamless tool expansion, automatic detection of new tools, and efficient, accurate reasoning even as tool numbers increase. Large Language Model (LLM)-based tool agents hold great potential, yet research that integrates physical sensing, dialogue, and multimodal emotional alignment within a unified framework is scarce, particularly in the context of Misty social robots.

\begin{figure}[t]
    \centering
    \includegraphics[width=1.0\linewidth]{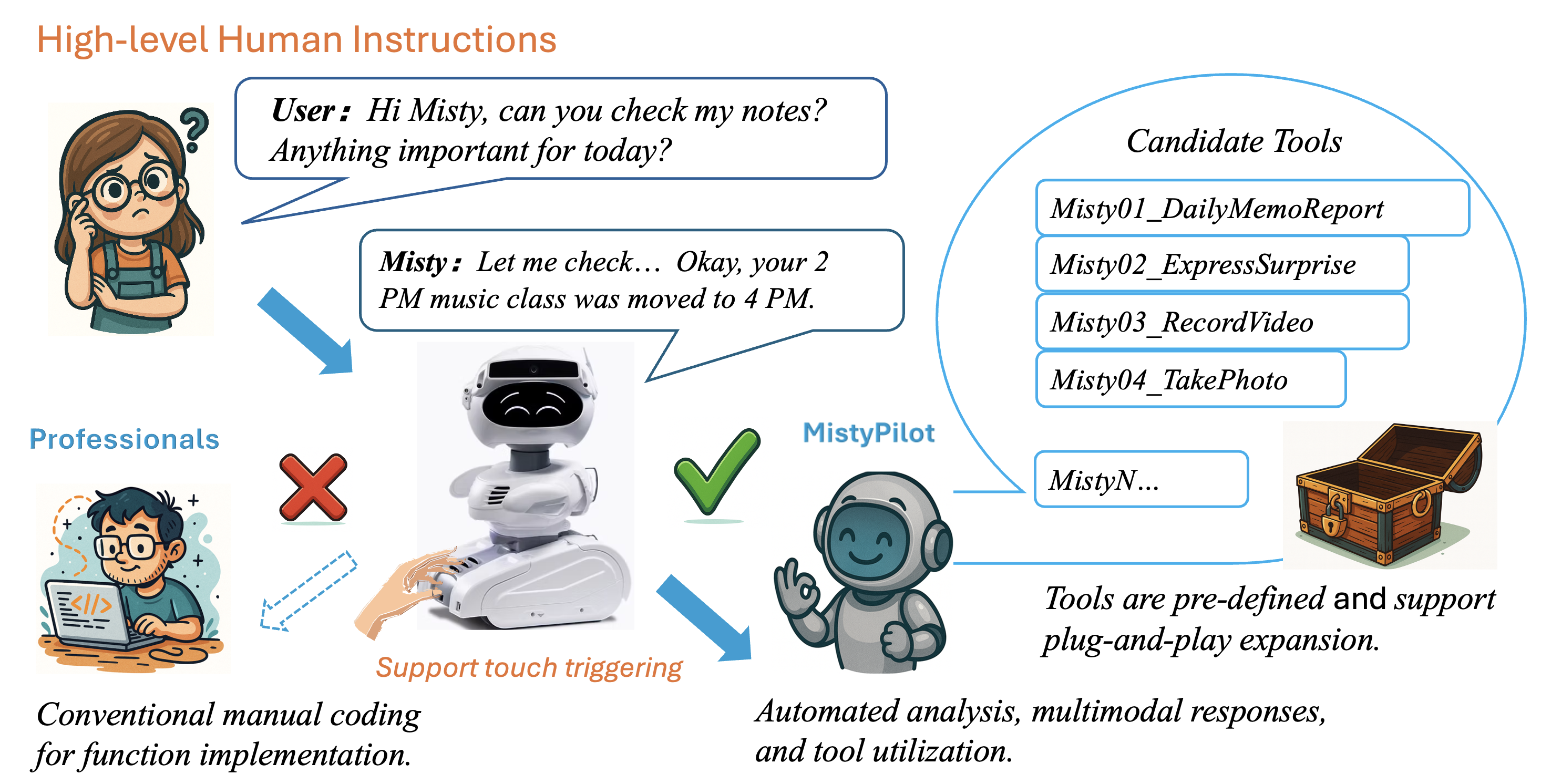}
    \caption{Overview of MistyPilot workflow for interpreting high-level human instructions. Instead of requiring professionals to hand-code robot behaviors and deploy features on the Misty robot, MistyPilot parses natural-language instructions, analyzes the task, selects and parameterizes tools from its library, and executes them on Misty Robot.}
    \label{fig:intrograph}
\end{figure}

Furthermore, most social robots continue to face significant challenges in achieving natural, emotionally attuned interactions. Many rely on basic text-to-speech (TTS) modules for voice generation and lack the ability to accurately perceive and align with emotional states, leading to responses that appear mechanical or emotionally detached. Moreover, these systems often fail to retain memory of prior interactions or adapt to user-specific preferences, resulting in repetitive reasoning and suboptimal interactions. This deficiency in emotional awareness and personalization greatly constrains their potential to serve as truly socially intelligent companions capable of fostering trust and sustaining long-term engagement.

To address these challenges, we propose MistyPilot, as illustrated in Fig.~\ref{fig:intrograph}, an agentic LLM-driven pilot framework that enables autonomous tool orchestration and parameter configuration, empowering social robots to deliver natural, engaging interactions and positioning them as more effective assistants and companions. 
To the best of our knowledge, this is the first agentic framework that unifies physical sensing, dialogue generation, and emotional alignment to enable autonomous tool orchestration for social robots, supported by advanced reasoning and decision-making autonomy. Our main contributions are summarized as follows:

\begin{itemize}
\item We present MistyPilot, an agentic LLM-driven framework for end-to-end interpretation of high-level user instructions, covering task analysis, tool selection, parameter configuration, and orchestration, while supporting plug-and-play tool expansion and maintaining system robustness.

\item We introduce a multi-agent architecture consisting of three components: a \textit{Task Router} that serves as the central dispatcher, a \textit{Physically Interactive Agent (PIA)} that handles dynamic sensor-triggered events and tool utilization, and a \textit{Socially Intelligent Agent (SIA)} that integrates a fast-slow thinking script-writing paradigm to achieve fine-grained multimodal emotional alignment in dialogue.

\item We contribute five benchmark datasets and conduct extensive experiments on evaluating task routing correctness, \textit{PIA} and \textit{SIA} performance, fast–slow thinking retrieval, and tool extensibility, demonstrating MistyPilot’s effectiveness and robustness. Human evaluation further suggests that MistyPilot is preferred in terms of response naturalness, accuracy, emotional expressiveness, timeliness, and overall user satisfaction. 
\end{itemize}

\begin{figure*}[t]
    \centering
    \includegraphics[width=\textwidth]{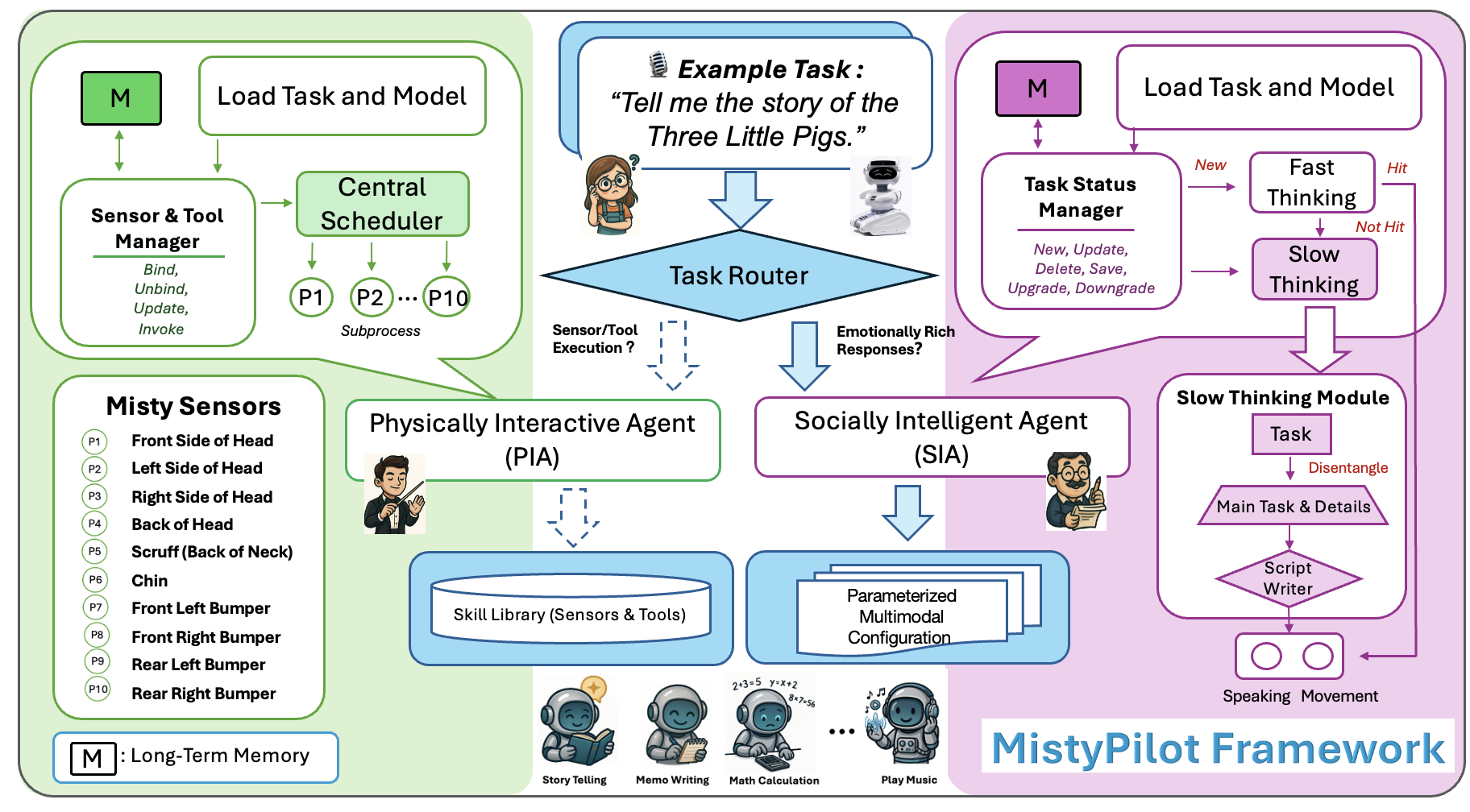}
    \caption{Overview of the MistyPilot framework. MistyPilot interprets high-level user instructions through a \textit{Task Router} that dispatches tasks to either a \textit{Physically Interactive Agent (PIA)}  or a \textit{Socially Intelligent Agent (SIA)}  for automated tool utilization and parameter adaptation. The \textit {PIA} oversees a \textit {Sensor \& Tool Manager} to dynamically orchestrate sensor-related and tool-dependent tasks, while the \textit {SIA} maintains a \textit{Task Status Manager} to track the current task state.
    \textit{Fast Thinking} accelerates inference by memory hitting for stored knowledge and preferences, while \textit{Slow Thinking} employs a \textit{Script Writer} to interpret the task, align responses with fine-grained emotional expression, and deliver them via speaking and movement modules.}
    \label{fig:overview}
\end{figure*}

\section{RELATED WORKS}
Recent advances in LLMs have demonstrated remarkable reasoning capabilities, motivating their integration into social robotics to enable end-to-end solutions for natural, high-level user interaction.
\subsection{Tool-augmented LLMs}
Recent advances in tool-augmented LLMs reflect a shift from solely enhancing reasoning performance to pursuing broader and more comprehensive capabilities. Toolformer \cite{schick2023toolformer} introduces a self-supervised approach that enables LLMs to autonomously learn API calls and embed them directly into text generation, while ReAct \cite{yao2023react} proposes interleaved reasoning and acting to dynamically query external tools and refine reasoning, effectively mitigating hallucination. ToolLLM \cite{qin2023toolllm} employs a tree-structured reasoning framework with rollback mechanisms to improve multi-tool orchestration, and Tulip Agent \cite{ocker2024tulip} proposes a vector store–based recursive tool retrieval system that scales to large tool libraries while reducing costs and supporting robotics applications.
 
Benchmark datasets such as API-Bank\cite{li2023api}, APIBench\cite{patil2024gorilla}, and ToolAlpaca \cite{tang2023toolalpaca} have further advanced this research by enabling standardized evaluation and generalization testing.
Building on these foundations, embodied and multimodal research
such as  AutoRT \cite{ahn2024autort}, Odyssey \cite{liu2024odyssey}, and Speech-enabled LLM Robot Control \cite{kadri2025llm}
further advanced tool-augmented LLM tasks by integrating perception and planning in real-world scenarios. However, comprehensive tool orchestration for social robot remains largely underexplored, highlighting the motivation for MistyPilot.

\subsection{Social Robots for Human–Robot Interaction}
Recent studies demonstrate the expanding role of social robots in providing companionship and emotional support across diverse scenarios. A recent survey \cite{kling2025social} summarize applications of social robots in adult psychiatry (e.g., depression, anxiety, loneliness), evaluating their potential to improve mental health and quality of life. 
However, the emotional expressiveness of social robots in interaction remains limited and requires further advancement. For older adults with dementia and depression, systems such as Ryan \cite{abdollahi2022artificial} integrate multimodal emotion recognition and affective dialogue management to enhance engagement and mood, yet speech synthesis remains largely neutral in prosody. AutoMisty \cite{wang2025automisty} introduces an LLM-driven framework for automated code generation to lower the barrier to robot customization, yet speech output still relies on standard, monotone TTS pipelines. 
In contrast, MistyPilot aims to bridge this gap by achieving multi-dimensional emotional alignment through adaptive motion, context-aware dialogue, and expressive speech synthesis, automatically adjusting tone, prosody, and arousal for more natural and empathetic interactions.

\section{PROBLEM FORMULATION}

To handle the open-ended interaction scenarios faced by social robots, we first formalize a Task Space $\mathcal{T}$. Given a natural language task instruction $T_i \in \mathcal{T}$ from this space, the objective is to construct an end-to-end tool orchestration pipeline that executes the specified task. To handle heterogeneous tasks, MistyPilot first employs a Task Router $R$ to dispatch the initial task $T_i$ to the appropriate tool agent $K^*$:
\begin{equation}
K^* = R(T_i), \quad K^* \in \{\textit{PIA, SIA}\}
\end{equation}

Once dispatched to the selected agent branch, the corresponding agent (SIA or PIA) parses the task $T_i$, automatically selects the required functions $f_j$ from a predefined tool library $\mathcal{F}$, and infers their corresponding parameter configurations $\theta_j \in \Theta_{f_j}$. Since completing a task typically requires invoking a series of functions, the final execution of task $T_i$ is represented as an ordered sequence of parameterized function calls:
\begin{equation}
\textit{Agent} (T_i) = \bigl( f_1(\theta_1), f_2(\theta_2), \dots, f_N(\theta_N) \bigr)
\end{equation}

Based on this formulation, the core objective of MistyPilot is to infer a valid sequence of function calls that satisfies the task requirements:
\begin{equation}
\textit{MistyPilot}(T_i) = \textit{Agent}_{K^*}(T_i) = \bigl( f_j(\theta_j) \bigr)_{j=1}^N
\end{equation}

\section{METHODOLOGY}

To handle open-ended human–robot interactions, MistyPilot adopts a multi-agent framework built on the tool agent paradigm, separating task routing, physical interaction, and social dialogue.
A tool agent \cite{wang2024tools, qu2025tool, li2025review} is characterized as an autonomous entity that translates high-level natural-language instructions into parameterized tool calls, invoking registered tools to execute complex tasks as shown in Fig.~\ref{fig:overview}.
Crucially, this multi-agent framework adopts an architectural-level role separation rather than mere prompt engineering. Unlike single-agent systems where context and all tools are fully collapsed, each agent in MistyPilot maintains its own localized state and tool-use space.
The following subsections describe the specific design of the \textit{Task Router}, \textit{SIA}, and \textit{PIA}.

\subsection{Task Router}
Given a task, MistyPilot first employs the \textit{Task Router}, a LLM agent, as the central dispatcher. It analyzes the high-level natural language instruction to determine the core intent of the user. 
If the task involves sensor-triggered events or direct tool invocation, the router dispatches it to the \textit{PIA}.
Conversely, if the task is dialogue-oriented, it is routed to the \textit{SIA} to generate an emotionally appropriate response.

\subsection{Social Intelligence Agent (SIA)}

When a dialogue-oriented task is dispatched to the \textit{SIA}, it is first processed by the \textit{Load Task and Model} module for state parsing and selection of the appropriate LLM version for reasoning. Subsequently, the \textit{Task Status Manager}, \textit{Fast Thinking}, and \textit{Slow Thinking }modules coordinate to orchestrate the necessary tools for completing the task.

\textbf{\textit{Load Task and Model}:}
This module converts user instructions into a task-state representation via LLM and a schema-constrained system prompt. It maps natural-language intent to the predefined action set in Table~1: \texttt{NEW}, \texttt{UPDATE}, and \texttt{DELETE} support task-state tracking and modification; \texttt{UPGRADE} and \texttt{DOWNGRADE} switch between a higher-capacity and a lightweight LLM for subsequent script generation; and \texttt{MEMORY} supports long-term information persistence upon user request.

\begin{table}[h]
    \caption{Keyword set $\mathcal{A}$ for task state parsing}
    \label{tab:task-parser}
\centering
\begin{tabular}{l|l}
\hline
Task State  &  System Control \\ \hline
\(\textbf{NEW}\langle \textit{main\_task}, \textit{details}\rangle\) & \(\textbf{UPGRADE}\) \\
\(\textbf{UPDATE}\langle \textit{details}\rangle\)                  & \(\textbf{DOWNGRADE}\) \\
\(\textbf{DELETE}\langle \textit{details}\rangle\)                   & \(\textbf{MEMORY}\) \\
\hline
\end{tabular}
\end{table}

\textbf{\textit{Task State Manager (TSM):}}
Prior studies show that LLMs accumulate faithfulness errors during iterative dialogue summarization, particularly in long context settings. \cite{liu2024lost,peysakhovich2023attention,maynez2020faithfulness,chang2023booookscore}.
To address this issue and stabilize behavior, we use an explicit, editable external memory to store and manage task states, reducing dependence on full context recomputation. This design ensures that each update is atomic and localized, without affecting other parts.
In Table \ref{tab:tsm_example}, We show an example of a user prompt and the TSM state. The details are split into semicolon-separated atomic units. When the user updates a detail (e.g., changing the time to 6 PM), only the relevant unit is modified and the rest remains unchanged.

\begin{table}[h]
\caption{An example task status managed by the TSM}
\label{tab:tsm_example}
\centering
\begin{tabular}{@{}p{0.15\columnwidth} >{\raggedright\arraybackslash}p{0.8\columnwidth}@{}}
\hline
User Task& \textcolor{blue}{“Hi Misty, I’d like to plan a day trip to New York City for tomorrow. Please create an itinerary that allows me to enjoy the city and return to my hotel by 7:00 PM.”} \\
\hline
 & \textbf{TSM} \\
\hline
\textbf{$\mathit{main\_task}$} & \textcolor{red}{Plan a day trip to New York City} \\
\textbf{$\mathit{details}$} & \textcolor{cyan!70!black}{Date tomorrow; return by 7:00 PM; goal enjoy the city}\\
\hline
\end{tabular}
\end{table}

\textbf{ \textit {Fast and Slow Thinking}: }
Inspired by fast and slow cognition in human problem solving \cite{kahneman2011thinking,pan2025survey,evans2008dual}, we adopt a dual-channel strategy: \textit{Fast Thinking} (Retrieve \& Reuse) as the default, and \textit {Slow Thinking} (Generate from Scratch) as a fallback.
In \textit{Fast Thinking}, we perform rapid retrieval over the \textbf{$\mathit{main\_task}$} content. We encode \textbf{$\mathit{main\_task}$} into a query embedding \textbf{$\mathbf{q}$} and compare it against the embeddings \textbf{$\mathbf{x}$} stored in local memory $\mathcal{M}$ by computing cosine distance, as defined in Eq.~\eqref{eq:threshold}. 
Here, the embedding space can be flexibly chosen. We compare three different embedding spaces, all of which demonstrate the robustness of our design (see Table~\ref{tab:test4}).
We set the threshold $\tau=0.4$ based on the observed distance distribution. If the minimum distance exceeds $\tau$, we treat the task as a retrieval miss and fall back to Slow Thinking for reasoning and generation.

\begin{equation}
    \min_{\mathbf{x} \in \mathcal{M}} d_{cos}(\mathbf{q}, \mathbf{x}) \le \tau
    \label{eq:threshold}
\end{equation}

In \textit{Slow Thinking}, the \textit{Script Writer} performs task orchestration through fine-grained emotional alignment at the utterance level.
Each utterance $u_i$ is aligned with a multimodal emotion label drawn from eight categories based on the extended Ekman emotion taxonomy~\cite{dalgleish2000handbook}: \textit{Happiness}, \textit{Sadness}, \textit{Anger}, \textit{Fear}, \textit{Disgust}, \textit{Surprise}, \textit{Contempt}, and \textit{Neutral}. Table~\ref{tab:script_segments} illustrates this process.

We express emotions consistently across three modalities: text, motion, and voice. Text is generated via context-aware reasoning. For motion, Misty executes expressive behaviors (e.g., arm waving, head tilting, body swaying, and light flashing) using templates aligned with Ekman’s basic emotions \cite{dalgleish2000handbook}, with LED colors and facial expressions reinforcing discriminability and cross-modal consistency. For voice, we bypass the built-in TTS and use OpenAI-TTS for controllable speech synthesis (timbre, speaking rate, and intonation). It (i) adapts prosody/intonation to text semantics and punctuation, (ii) injects emotion-conditioned style by encoding affective attributes (e.g., arousal, intensity, prosodic patterns) \cite{goudbeek2010beyond}, and (iii) controls speaking rate conditioned on emotion labels. Following \cite{goudbeek2010beyond}, we discretize speaking rate by arousal: $1.00$ (baseline), $0.95$ (low arousal), and $1.05$ (high arousal), yielding synchronized action--speech expressions. 
Finally, the \textit {SIA} delivers motion and voice to the \textit {Speaking} and \textit {Movement} modules for simultaneous execution on Misty.

\begin{table}[h]
\centering
\caption{Emotional attunement example by the \textit{Script Writer}}
\label{tab:script_segments}
\begin{tabular}{@{}p{0.1\columnwidth} p{0.6\columnwidth} p{0.20\columnwidth}@{}}
\toprule
\textbf{Prompt} & \textcolor{blue}{"Tell me the story of Three Little Pig"} &  \\
\toprule
\textbf{Utterance} & \textbf{Text} & \textbf{Emotion} \\
\midrule
1 & Three little pigs left home to build their own houses. & Neutral \\
2 & The first pig quickly built a house of straw. & Contempt \\
3 & The second pig put up a house of sticks with little effort. & Contempt \\
4 & The third pig worked diligently to lay strong bricks for a sturdy home. & Happiness \\
\midrule
\multicolumn{3}{c}{\dots} \\
\bottomrule
\end{tabular}
\end{table}

\subsection{Physically Interactive Agent (PIA)}

Tasks involving sensor-triggered events or direct tool invocation are routed to \textit{PIA}. In human--robot interaction, \textit{PIA} uses the \textit{Load Task and Model} module to interpret user intent and either (i) directly execute a skill tool or (ii) identify the target sensor and retrieve the corresponding tool from the \textit{Skill Library}. Subsequently, \textit{PIA} leverages the \textit{Sensor \& Tool Manager} module to complete sensor bindings and execute the tool, enabling seamless human-Misty interaction.

\textbf{\textit{Load Task and Model}:}
Similarly, this module converts user instructions into sensor- or tool-related commands via LLM and a schema-constrained system prompt. It maps natural-language intent to the predefined command set in Table~\ref{tab:bind_commands}: \texttt{BIND}, \texttt{UPDATE}, and \texttt{UNBIND} support customized sensor interactions, while \texttt{INVOKE}
enables direct one-time tool execution without sensor binding.

\begin{table}[h]
\centering
\caption{Command Set for Sensor \& Tool Management
 }
\label{tab:bind_commands}
\begin{tabular}{c|l}
\toprule
\textbf{Command} & \textbf{Description} \\
\midrule
\texttt{ \textbf{BIND}$\langle s, k \rangle$}   & Bind skill $k$ to sensor $s$. \\
\texttt{ \textbf{UPDATE}$\langle s, k \rangle$} & Update skill $k$ to sensor. $s$ \\
\texttt{\textbf{UNBIND}$\langle s \rangle$}    & Unbind sensor $s$ from all associated skills. \\
\texttt{\textbf{INVOKE}$\langle k \rangle$}    & Directly invoke skill $k$. \\
\bottomrule
\end{tabular}
\end{table}

In Table~\ref{tab:bind_commands}, $s \in S$ denotes a valid sensor ID from the system sensor namespace $S$ (Misty has ten sensors; see Fig.~\ref{fig:overview}, \textit{Misty Sensors}), and $k \in K$ denotes a callable general-purpose skill tool (e.g., ``play relaxing piano music''). The skill set $K$ is loaded at runtime, enabling plug-and-play tool expansion: at startup, MistyPilot scans the local skill directory, parses docstrings to build an inventory (function name and capability description), and injects it into the \textit{PIA} system prompt.

\textbf{\textit{Sensor \& Tool Manager (STM)}:}
\textit{STM}, through a \textit{Central Scheduler} module, maintains an independent worker process for each valid sensor and records their status in a process table (\textit{sensor}, \textit{PID},  \textit{bound skill}, \textit{Status}).  
The \textit{Central Scheduler} performs periodic active checks, and any inactive process is immediately restarted, ensuring robustness and availability. 
Table~\ref{tab:process_table} presents example entries of the \textit{STM} process table. The fields \textit{sensor}, \textit{PID}, and \textit{bound skill} denote the sensor name, process identifier, and the associated skill, respectively. The \textit{Status} field indicates whether a sensor’s PID exists in the system process tree (\textit{Active}) or has terminated unexpectedly (\textit{Inactive}).

Upon MistyPilot startup, \textit{STM} scans the persisted process table, recovers the sensor-skill bindings, and automatically remounts them on Misty. This mechanism ensures persistent state and rapid recovery across sessions.

\begin{table}[h]
\centering
\caption{Example of STM process table entries}
\label{tab:process_table}
\begin{tabular}{p{0.25\columnwidth}|p{0.15\columnwidth}|p{0.25\columnwidth}|p{0.1\columnwidth}}
\toprule
\textbf{Sensor} & \textbf{PID} & \textbf{Bound Skill} & \textbf{Status} \\
\midrule
Head Touch        & 1023 & Greeting        & \textcolor{darkgreen}{Active} \\
Scruff Touch      & 1098 & DailyMemoReport & \textcolor{darkgreen}{Active} \\
Front Left Bumper & 1120 & NodHead         & \textcolor{darkred}{Inactive} \\
\bottomrule
\end{tabular}
\end{table}

\section{Benchmark Dataset}

To comprehensively evaluate the MistyPilot framework, including the rationality of its modular design, robustness, generalization, and extensibility, we curate five benchmark datasets. {The benchmark construction follows a hybrid pipeline \cite{wang2023self}} designed to synthesize realistic instructions while mitigating single-model bias: (1) we collect empirically grounded seeds from experts with substantial on-site Misty demonstration experience; (2) we expand these seeds using a multi-LLM ensemble (Gemini 2.5 Pro and GPT-5 ) under explicitly defined Easy/Hard criteria, with single-feature iterative prompting to control complexity; (3) we enforce diversity via strict ROUGE-L filtering, removing samples with overlap greater than 0.7 to reduce redundancy; (4) we randomly sample from the instruction pool to form the final dataset and conduct structured annotation; and (5) we perform human-in-the-loop annotation to assign deterministic ground-truth routing decisions, target functions, and parameter specifications, 
enabling quantitative evaluation of MistyPilot’s decision accuracy.

\textbf{MistyPilot-Route100}: This dataset comprises 100 task instructions designed to evaluate the routing capability of MistyPilot, specifically whether a task should be dispatched to the \textit{SIA} for dialogue-oriented task or to the \textit{PIA} for sensor-triggered events or direct tool invocation. The dataset is relatively balanced across both categories, with 58 \textit{SIA} tasks and 42 \textit{PIA} tasks. Each subset is further divided into \emph{Easy} cases (direct and straightforward instructions) and \emph{Hard} cases (implicit or composite conditions), enabling a fine-grained evaluation of routing robustness under diverse natural-language inputs.

\textbf{{MistyPilot-SensorBind40}}: This dataset contains 40 single-turn instances spanning two categories: (i) sensor-binding commands and (ii) immediate single-tool invocation commands without sensor binding, curated to evaluate the \textit{PIA’s} routing accuracy and sensor-binding capability. The instances are split by difficulty: 20 \emph{Easy} cases that require either a single, immediate sensor-grounded response or the immediate invocation of a single tool (e.g., ``touch your head and make a cute sound'', or ``I'm doing home exercise, play some good workout music for me''), and 20 \emph{Hard} cases that require parsing and registering multiple concurrent sensor-trigger bindings within a single pass (e.g., ``when I tap your chin, take a photo; press your forehead to say hi; touch your right side to show sadness''). These multi-binding instructions are challenging because all sensor bindings must be correctly parsed, registered, and executed without omission.

\textbf{MistyPilot-TaskParser256}:This dataset contains 40 multi-turn dialogues (256 turns total) spanning multiple domains, including daily-life assistance, planning, storytelling, and emotion supportive interaction. It is designed to evaluate the \textit{SIA}'s proficiency in parsing user intent, dynamically managing task states and executing system-level controls (e.g., switching to a more powerful model) based on explicit user feedback.The dataset is split into 7 \emph{Easy} dialogues (28 turns) and 33 \emph{Hard} dialogues (228 turns). \emph{Easy} dialogues contain clear instructions with no more than four turns, while \emph{Hard} dialogues involve longer interactions with ambiguous references (coreference), topic shifts, and interleaved or evolving instructions, posing challenges for long-horizon reasoning and context-aware adaptation.
 
\textbf{MistyPilot-FastThinking230}: This dataset targets MistyPilot’s \emph{fast-thinking} module and evaluates its retrieval accuracy and robustness to paraphrasing. It contains 230 commands derived from 46 canonical tasks, each expanded into five variants that preserve the same core task while varying surface details. The goal is to measure whether the system can correctly retrieve and reuse previously executed implementations under surface-level linguistic variations, thereby reducing latency and computational cost.
  
\textbf{MistyPilot ToolExtension30/50/70/100}: This data set evaluates how well MistyPilot scales to newly introduced tools, assessing \textit{PIA} extensibility, generalization and adaptability under dynamically injected skills. New tools are introduced at four scales (30, 50, 70, and 100 tools), and each tool is provided solely by its docstring; the system is expected to select the appropriate tool based only on these descriptions.

\section{Experiments}

To comprehensively evaluate MistyPilot, we conduct experiments on \textit{Task Routing} Correctness, \textit{PIA} Performance, \textit{SIA} Performance, Fast-Thinking Retrieval, and Tool Extensibility using their respective datasets. All other settings are kept identical (GPT-5-mini backbone), with each configuration executed five times and results reported as mean ± standard deviation.
MistyPilot adopts an architectural-level role separation multi-agent design, where the contribution lies in system architecture rather than prompt engineering. The \textit{Task Router}, \textit{PIA}, and \textit{SIA} operate under strict logical and execution isolation, each maintaining its own context and restricted tools. To validate this design, we also construct a Single-Agent baseline that collapses \textit{Task Router}, \textit{PIA}, \textit{SIA}, and all tools into a single LLM agent, requiring unified end-to-end task interpretation and tool execution within one control flow.

\subsection{Evaluation of Task Routing Correctness}
 
In this section, we evaluate MistyPilot’s \textit{Task Router} by comparing its routing accuracy across different task categories and difficulty levels on the MistyPilot-Route100 benchmark dataset.
The ability of the \textit{Task Router} to correctly identify the appropriate agent is the most critical first step in task execution. When semantically overlapping or closely related concepts cause tool-selection ambiguity, even a minor routing error can directly result in task failure.

To evaluate the effectiveness of our Multi-Agent design for task routing, we conduct comparative experiments against a Single-Agent baseline.
As shown in Table~\ref{tab:task1}, our Multi-Agent design consistently outperforms the baseline, achieving 100\% accuracy on SIA–Easy/Hard and PIA–Easy, and delivering higher and more stable performance on PIA–Hard (96.2\%$\pm$5.66\% vs.\ 90.4\%$\pm$15.65\%).
These findings indicate that architectural-level separation in MistyPilot contributes to reduced routing ambiguity and more robust routing behavior.

\begin{table}[t]
\centering
\caption{Task Routing Correctness on MistyPilot-Route100 Dataset.(mean$\pm$std over 5 runs).}
\setlength{\tabcolsep}{6pt}
\renewcommand{\arraystretch}{1.2}
\begin{tabular}{lcc}
\toprule
\textbf{Condition} & \textbf{MistyPilot (Multi-Agent)} & \textbf{Single-Agent} \\
\midrule
SIA -- Easy  & $100\% \pm 0.00\%$     & $99.2\% \pm 1.79\%$ \\
SIA -- Hard  & $100\% \pm 0.00\%$     & $100\% \pm 0.00\%$ \\
PIA -- Easy  & $100\% \pm 0.00\%$     & $100\% \pm 0.00\%$ \\
PIA -- Hard  & $96.2\% \pm 5.66\%$ & $90.4\% \pm 15.65\%$ \\
\bottomrule
\end{tabular}
\label{tab:task1}
\end{table}

\subsection{Evaluation of PIA Performance}

The evaluation of \textit{PIA} focuses on whether dynamic sensor–tool bindings are correctly established and whether tool invocations without sensor binding are executed properly.
We conduct experiments on the MistyPilot-SensorBind40 dataset and follow the same ablation setting as above, comparing the Multi-Agent design with the Single-Agent baseline.

The experimental results, as shown in Table~\ref{tab:task2}, indicate that both designs achieve 100\% accuracy on the Easy subset. On the Hard subset, however, our MistyPilot (Multi-Agent) design performs  better and  more stably, reaching 100\%\,$\pm$\,0\% accuracy compared to  81.00\%\,$\pm$\,4.18\% for the Single-Agent baseline. 
Overall, MistyPilot’s design enhances both correctness and stability in sensor–tool binding and sensor-free tool invocation.

\begin{table}[t]
  \centering
  \caption{PIA Performance Evaluation on the MistyPilot-SensorBind40 Dataset (mean$\pm$std over 5 runs).}
  \setlength{\tabcolsep}{6pt}
  \renewcommand{\arraystretch}{1.2}
  \begin{tabular}{lcc}
    \toprule
    \textbf{Subset} & \textbf{MistyPilot (Multi-Agent)} & \textbf{Single-Agent} \\
    \midrule
    Easy & $100.00\% \pm 0.00\%$ & $100.00\% \pm 0.00\%$ \\
    Hard & $100.00\% \pm 0.00\%$  & $81.00\% \pm 4.18\%$  \\
    \bottomrule
  \end{tabular}
  \label{tab:task2}
\end{table}

\begin{table}[t]
  \centering
  \caption{SIA Performance Evaluation on MistyPilot-TaskParser256 Dataset (mean$\pm$std over 5 runs).}
  \setlength{\tabcolsep}{6pt}
  \renewcommand{\arraystretch}{1.2}
  \begin{tabular}{lcc}
    \toprule
    \textbf{Subset} & \textbf{MistyPilot (Multi-Agent)} & \textbf{Single-Agent} \\
    \midrule
    Easy & $99.29\% \pm 1.60\%$ & $91.43\% \pm 13.27\%$ \\
    Hard & $96.75\% \pm 1.15\%$ & $93.50\% \pm 5.36\%$  \\
    \bottomrule
  \end{tabular}
  \label{tab:task3}
\end{table}

\subsection{Evaluation of SIA Performance}

For \textit{SIA}, the \textit{Task Status} is critical in multi-turn dialogues, as accurate status recognition (e.g., \texttt{UPDATE} vs.\ \texttt{DELETE}) directly affects downstream module coordination.Moreover, the correctness of \textit{System Control} actions also has a direct impact on the overall system behavior and performance. We evaluate their correctness on MistyPilot-TaskParser256, comparing MistyPilot (Multi-Agent) with a Single-Agent baseline.

As shown in Table~\ref{tab:task3}, MistyPilot achieves higher accuracy with markedly lower variance on both subsets: 99.29\%\,$\pm$\,1.60\% vs.\ 91.43\%\,$\pm$\,13.27\% on Easy, and 96.75\%\,$\pm$\,1.15\% vs.\ 93.50\%\,$\pm$\,5.36\% on Hard. 
Overall, MistyPilot’s \textit{SIA} demonstrates more robust behavior in long-horizon interactions involving ambiguous references (e.g., coreference), topic shifts, and evolving instructions.

\subsection{Evaluation of Fast-Thinking Retrieval}

Fast Thinking reuses previously generated results that users find useful. For similar requests, it retrieves instead of regenerating from scratch, reducing redundant computation and improving response speed. We evaluate this on the MistyPilot-FastThinking230 dataset.
MistyPilot disentangles the task status into \textit{main\_task} and \textit{details}, and performs fast-thinking retrieval using the \textit{main\_task} representation as \textit{Fast Path}. We compare this strategy with a \textit{Raw Text} (original input) baseline as an ablation study. 
We evaluate both settings using three metrics: \textit{Top-1 Accuracy}, \textit{Rank1 Dist. Mean}, and \textit{Rank2 Dist. Mean}, where the latter two denote the average distance between the query and its top-1 and top-2 candidates in the embedding space, respectively. 
We additionally evaluate our method across three embeddings, two proprietary embedding (text-embedding-3-large and text-embedding-3-small) and one open-source embedding (all-MiniLM-L6-v2), to ensure robustness across different embedding spaces.

The experimental results, as shown in Table~\ref{tab:test4}, demonstrate that using MistyPilot’s disentangled representation achieves 100.00\% Top-1 accuracy across all three embeddings, whereas the raw-text baseline only reaches 67.83\%, 72.61\%, and 58.70\%, respectively.
Meanwhile, the margin between \textit{Rank2 Dist. Mean} and \textit{Rank1 Dist. Mean} under Misty’s disentangled representation is substantially larger than that of the  \textit{Raw Text} baseline, indicating stronger discriminability and a clearer separation between the correct answer and the second-best candidate. Overall, Misty’s disentangled design achieves superior fast-thinking performance across both open-source and proprietary embeddings.

In addition, to better illustrate the response time difference between the fast-thinking and slow-thinking paths, we conduct a latency comparison experiment. The results show that the fast-thinking path significantly reduces the response time from $5.088 \pm 2.571$s to $2.263 \pm 0.627$s, corresponding to a 55.5\% reduction.

\begin{table}[t]
  \centering
  \caption{Fast-Thinking Retrieval Evaluation on MistyPilot-FastThinking230 Dataset: \textit{Raw text} vs.\ \textit{Fast Path}.}
  \label{tab:test4}
  \begingroup
  \setlength{\tabcolsep}{4pt} 
  \renewcommand{\arraystretch}{1.2}
  \small

  \begin{threeparttable}
  \begin{tabular}{lllll}
    \toprule
    \textbf{Data} & \textbf{Model} & \textbf{Top-1 (\%)} & 
    \textbf{Rank1 Dist.} & \textbf{Rank2 Dist.} \\
    & & & \textbf{Mean (Std)} & \textbf{Mean (Std)} \\
    \midrule
    Raw Text  & TE3-L  & 67.83\%  & 0.365 (0.070) & 0.464 (0.071) \\
    Raw Text  & TE3-S  & 72.61\%  & 0.342 (0.082) & 0.453 (0.066) \\
    Raw Text  & MiniLM & 58.70\%  & 0.397 (0.127) & 0.570 (0.066) \\
    \midrule
    Fast Path & TE3-L  & 100.00\% & 0.003 (0.015) & 0.392 (0.216) \\
    Fast Path & TE3-S  & 100.00\% & 0.003 (0.015) & 0.394 (0.211) \\
    Fast Path & MiniLM & 100.00\% & 0.000 (0.002) & 0.463 (0.248) \\
    \bottomrule
  \end{tabular}

  \begin{tablenotes}[flushleft]
    \footnotesize
    \item \textit{Note:} TE3-L = text-embedding-3-large; TE3-S = text-embedding-3-small; MiniLM = all-minilm-l6-v2.
  \end{tablenotes}
  \end{threeparttable}
  \endgroup
\end{table}

\subsection{Evaluation of Tool Extensibility}
We aim to build an end-to-end, plug-and-play tool agent framework; therefore, this part of the experiments focuses on evaluating tool extensibility and scalability. Our experiments are conducted on the MistyPilot-ToolExtension30/50/70/100 dataset. We expand the skill set to 30, 50, 70 and 100 tools, and repeat each configuration five times.

The results, shown in Table~\ref{tab:test6}, indicate that our proposed design achieves 100.00\% success rate in all runs at every scale, demonstrating the strong extensibility of the dynamic injection mechanism.

\begin{table}[t]
  \centering
  \caption{Tool Extensibility Evaluation on MistyPilot-ToolExtension30/50/70/100 Dataset}
  \label{tab:test6}
  \setlength{\tabcolsep}{6pt}
  \renewcommand{\arraystretch}{1.2}
  \begin{tabular}{@{}lcccc@{}}
    \toprule
    \textbf{Scale} & \textbf{30} & \textbf{50} & \textbf{70} & \textbf{100} \\
    \midrule
    Success rate & 100.00\% & 100.00\% & 100.00\% & 100.00\% \\
    \bottomrule
  \end{tabular}
\end{table}

\subsection{Human-Robot Interaction Evaluation}

To more comprehensively evaluate our system, we conducted a human–robot interaction study with 12 volunteers. Each participant interacted one-on-one with the Misty robot powered by MistyPilot for at least 30 minutes, totaling over 360 minutes (6 hours) of real-world interaction.
Participants engaged in open-domain conversations, receiving multimodal emotionally attuned responses via the \textit{SIA} module (e.g., “Could you tell me a ghost story?”), and invoked 30 utility tools via the \textit{PIA} module (e.g., weather queries, local recommendations, music playback, and photo capture). After the interaction, they completed a five-item post-study questionnaire (Table~\ref{tab:user_study}) assessing key aspects of their experience. This study was conducted under the approval of our institution’s Institutional Review Board (IRB).

This questionnaire was adapted from representative items in established HRI and UX instruments, including the Godspeed Questionnaire Series~\cite{bartneck2009measurement} and the User Experience Questionnaire (UEQ)~\cite{laugwitz2008construction}, and was further tailored to our multimodal interaction setting. All items were rated on a 5-point Likert scale (1 = strongly disagree'', 5 = strongly agree''). The five items cover Interaction Naturalness, Comprehension Accuracy, Emotional Expressiveness, System Responsiveness, and Overall User Satisfaction.
 
Statistical analysis of the post-study questionnaire indicates consistently positive feedback across all evaluation dimensions, with all mean scores exceeding 4.4/5.
Overall satisfaction is high ($Mean = 4.75$, $Std = 0.45$). The highest ratings are observed for Interaction Naturalness and Emotional Expressiveness (both $Mean = 4.75$), suggesting that MistyPilot supports natural and emotionally attuned interaction. Participants also report strong responsiveness ($Mean = 4.67$, $Std = 0.49$) and reliable comprehension accuracy ($Mean = 4.42$, $Std = 0.67$), indicating robust intent understanding under diverse natural-language inputs. 
We also observed occasional failure cases mainly due to automatic speech recognition (ASR) errors for accented speech, which sometimes led to intent misunderstanding and suboptimal responses.

\begin{table}[h]
\centering
\renewcommand{\arraystretch}{1.4}
\caption{Subjective Evaluation Results of the MistyPilot System }
\begin{tabular}{@{} p{0.75\linewidth} c c @{}}
\hline
\textbf{Evaluation Dimension and Questionnaire Item} & \textbf{Mean} & \textbf{Std} \\
\hline
\textbf{Interaction Naturalness:} Communicating with the robot through natural language to perform specific tasks was intuitive and straightforward. & 4.75 & 0.45 \\
\hline
\textbf{Comprehension Accuracy:} The robot accurately understood my intentions and provided responses that met my expectations. & 4.42 & 0.67 \\
\hline
\textbf{Emotional Expressiveness:} The robot's speech and reactions were emotionally expressive, giving it a highly anthropomorphic and human-like presence. & 4.75 & 0.45 \\
\hline
\textbf{System Responsiveness:} The overall response latency was within an acceptable range. & 4.67 & 0.49 \\
\hline
\textbf{User Satisfaction:} Overall, I am satisfied with the multimodal interaction experience provided by the MistyPilot system. & 4.75 & 0.45 \\
\hline
\end{tabular}
\label{tab:user_study}
\vspace{1ex} \\
\raggedright \footnotesize \textit{Note: All scores are reported on a 5-point Likert scale (1 = strongly disagree, 5 = strongly agree). Mean indicates the average rating, and Std indicates the standard deviation.}

\end{table}

\section{CONCLUSION}

We present MistyPilot, a multi-agent LLM-based framework that enables real-time spoken interaction to drive task execution and translates high-level user instructions into reliable robot behaviors. By automatically interpreting intent, selecting and configuring appropriate tools, and executing them robustly, MistyPilot addresses the core challenge of non-expert human–robot interaction.
This capability is achieved through a role-specialized architecture: the \textit{PIA} manages sensor–tool binding and sensor-free tool invocation, while the \textit{SIA} focuses on socially intelligent dialogue with emotional alignment. Furthermore, a fast–slow thinking paradigm enhances reasoning efficiency, and a plug-and-play design enables seamless tool extension. Extensive experiments validate its effectiveness in task routing, sentiment alignment, parameter configuration, and dynamic tool invocation, demonstrating robustness and generalizability. Human evaluation further indicates user preference for MistyPilot’s emotion-aligned multimodal responses.

Future work will focus on developing a holistic, co-optimized pipeline for tool generation and utilization, ultimately enabling self-evolving, continuously learning social robotic systems capable of delivering richer, context-aware, and human-centric interactions.

\scriptsize
\bibliographystyle{IEEEtran} %
\bibliography{IEEEexample} %

\begin{thebibliography}{10}
\providecommand{\url}[1]{#1}
\csname url@rmstyle\endcsname
\providecommand{\newblock}{\relax}
\providecommand{\bibinfo}[2]{#2}
\providecommand\BIBentrySTDinterwordspacing{\spaceskip=0pt\relax}
\providecommand\BIBentryALTinterwordstretchfactor{4}
\providecommand\BIBentryALTinterwordspacing{\spaceskip=\fontdimen2\font plus
\BIBentryALTinterwordstretchfactor\fontdimen3\font minus \fontdimen4\font\relax}
\providecommand\BIBforeignlanguage[2]{{%
\expandafter\ifx\csname l@#1\endcsname\relax
\typeout{** WARNING: IEEEtran.bst: No hyphenation pattern has been}%
\typeout{** loaded for the language `#1'. Using the pattern for}%
\typeout{** the default language instead.}%
\else
\language=\csname l@#1\endcsname
\fi
#2}}

\bibitem{wang2025social}
X.~Wang, Y.~Wang, Z.~Han, Z.~Duan, Z.~Zhang, and T.~A. Alsudais, ``Social robots for child development: research hotspots, topic modeling, and collaborations,'' \emph{Humanities and Social Sciences Communications}, vol.~12, no.~1, pp. 1--15, 2025.

\bibitem{jackson2009teachers}
J.~N. Jackson and J.~M. Campbell, ``Teachers’ peer buddy selections for children with autism: Social characteristics and relationship with peer nominations,'' \emph{Journal of Autism and Developmental Disorders}, vol.~39, no.~2, pp. 269--277, 2009.

\bibitem{gonzalez2025social}
P.~Gonz{\'a}lez-Oliveras, O.~Engwall, and A.~Wilde, ``Social educational robotics and learning analytics: A scoping review of an emerging field,'' \emph{International Journal of Social Robotics}, pp. 1--16, 2025.

\bibitem{romero2024using}
A.~J. Romero-C.~de Vaca, R.~A. Melendez-Armenta, and H.~Ponce, ``Using social robotics to identify educational behavior: A survey,'' \emph{Electronics}, vol.~13, no.~19, p. 3956, 2024.

\bibitem{sawik2023robots}
B.~Sawik, S.~Tobis, E.~Baum, A.~Suwalska, S.~Kropi{\'n}ska, K.~Stachnik, E.~P{\'e}rez-Bernabeu, M.~Cildoz, A.~Agustin, and K.~Wieczorowska-Tobis, ``Robots for elderly care: review, multi-criteria optimization model and qualitative case study,'' in \emph{Healthcare}, vol.~11, no.~9.\hskip 1em plus 0.5em minus 0.4em\relax MDPI, 2023, p. 1286.

\bibitem{zhao2025social}
I.~Y. Zhao, A.~Y.~M. Leung, Y.~Huang, and Y.~Liu, ``A social robot in home care: Acceptability and utility among community-dwelling older adults,'' \emph{Innovation in Aging}, vol.~9, no.~5, 2025.

\bibitem{wang2025automisty}
X.~Wang, L.~Dong, S.~Rangasrinivasan, I.~Nwogu, S.~Setlur, and V.~Govindaraju, ``Automisty: a multi-agent llm framework for automated code generation in the misty social robot,'' in \emph{2025 IEEE/RSJ International Conference on Intelligent Robots and Systems (IROS)}.\hskip 1em plus 0.5em minus 0.4em\relax IEEE, 2025, pp. 9194--9201.

\bibitem{ciuffreda2025design}
I.~Ciuffreda, G.~Amabili, S.~Casaccia, M.~Benadduci, A.~Margaritini, E.~Maranesi, F.~Marconi, A.~De~Masi, J.~Alberts, J.~de~Koning, \emph{et~al.}, ``Design and development of a technological platform based on a sensorized social robot for supporting older adults and caregivers: Guardian ecosystem,'' \emph{International Journal of Social Robotics}, vol.~17, no.~5, pp. 803--822, 2025.

\bibitem{eckert2025programming}
A.~Eckert and P.~Juvonen, ``Programming social robots in linguistically heterogeneous classrooms: the case of misty,'' \emph{Digital Experiences in Mathematics Education}, pp. 1--20, 2025.

\bibitem{schick2023toolformer}
T.~Schick, J.~Dwivedi-Yu, R.~Dess{\`\i}, R.~Raileanu, M.~Lomeli, E.~Hambro, L.~Zettlemoyer, N.~Cancedda, and T.~Scialom, ``Toolformer: Language models can teach themselves to use tools,'' \emph{Advances in Neural Information Processing Systems}, vol.~36, pp. 68\,539--68\,551, 2023.

\bibitem{yao2023react}
S.~Yao, J.~Zhao, D.~Yu, N.~Du, I.~Shafran, K.~Narasimhan, and Y.~Cao, ``React: Synergizing reasoning and acting in language models,'' in \emph{International Conference on Learning Representations (ICLR)}, 2023.

\bibitem{qin2023toolllm}
Y.~Qin, S.~Liang, Y.~Ye, K.~Zhu, L.~Yan, Y.~Lu, Y.~Lin, X.~Cong, X.~Tang, B.~Qian, \emph{et~al.}, ``Toolllm: Facilitating large language models to master 16000+ real-world apis,'' \emph{arXiv preprint arXiv:2307.16789}, 2023.

\bibitem{ocker2024tulip}
F.~Ocker, D.~Tanneberg, J.~Eggert, and M.~Gienger, ``Tulip agent--enabling llm-based agents to solve tasks using large tool libraries,'' \emph{arXiv preprint arXiv:2407.21778}, 2024.

\bibitem{li2023api}
M.~Li, Y.~Zhao, B.~Yu, F.~Song, H.~Li, H.~Yu, Z.~Li, F.~Huang, and Y.~Li, ``Api-bank: A comprehensive benchmark for tool-augmented llms,'' in \emph{Proceedings of the 2023 conference on empirical methods in natural language processing}, 2023, pp. 3102--3116.

\bibitem{patil2024gorilla}
S.~G. Patil, T.~Zhang, X.~Wang, and J.~E. Gonzalez, ``Gorilla: Large language model connected with massive apis,'' \emph{Advances in Neural Information Processing Systems}, vol.~37, pp. 126\,544--126\,565, 2024.

\bibitem{tang2023toolalpaca}
Q.~Tang, Z.~Deng, H.~Lin, X.~Han, Q.~Liang, B.~Cao, and L.~Sun, ``Toolalpaca: Generalized tool learning for language models with 3000 simulated cases,'' \emph{arXiv preprint arXiv:2306.05301}, 2023.

\bibitem{ahn2024autort}
M.~Ahn, D.~Dwibedi, C.~Finn, M.~G. Arenas, K.~Gopalakrishnan, K.~Hausman, B.~Ichter, A.~Irpan, N.~Joshi, R.~Julian, \emph{et~al.}, ``Autort: Embodied foundation models for large scale orchestration of robotic agents,'' \emph{arXiv preprint arXiv:2401.12963}, 2024.

\bibitem{liu2024odyssey}
S.~Liu, Y.~Li, K.~Zhang, Z.~Cui, W.~Fang, Y.~Zheng, T.~Zheng, and M.~Song, ``Odyssey: Empowering minecraft agents with open-world skills,'' \emph{arXiv preprint arXiv:2407.15325}, 2024.

\bibitem{kadri2025llm}
I.~Kadri, S.~A. Selouani, M.~Ghribi, R.~Ghali, and S.~Mekhoukh, ``Llm-driven agent for speech-enabled control of industrial robots: A case study in snow-crab quality inspection,'' \emph{Results in Engineering}, p. 106660, 2025.

\bibitem{kling2025social}
M.~Kling, A.~Haeussl, N.~Dalkner, F.~T. Fellendorf, M.~Lenger, A.~Finner, J.~Ilic, I.~S. Smolak, L.~Stojec, I.~Zwigl, \emph{et~al.}, ``Social robots in adult psychiatry: a summary of utilisation and impact,'' \emph{Frontiers in psychiatry}, vol.~16, p. 1506776, 2025.

\bibitem{abdollahi2022artificial}
H.~Abdollahi, M.~H. Mahoor, R.~Zandie, J.~Siewierski, and S.~H. Qualls, ``Artificial emotional intelligence in socially assistive robots for older adults: a pilot study,'' \emph{IEEE Transactions on Affective Computing}, vol.~14, no.~3, pp. 2020--2032, 2022.

\bibitem{wang2024tools}
Z.~Wang, Z.~Cheng, H.~Zhu, D.~Fried, and G.~Neubig, ``What are tools anyway? a survey from the language model perspective,'' \emph{arXiv preprint arXiv:2403.15452}, 2024.

\bibitem{qu2025tool}
C.~Qu, S.~Dai, X.~Wei, H.~Cai, S.~Wang, D.~Yin, J.~Xu, and J.-R. Wen, ``Tool learning with large language models: A survey,'' \emph{Frontiers of Computer Science}, vol.~19, no.~8, p. 198343, 2025.

\bibitem{li2025review}
X.~Li, ``A review of prominent paradigms for llm-based agents: Tool use, planning (including rag), and feedback learning,'' in \emph{Proceedings of the 31st International Conference on Computational Linguistics}, 2025, pp. 9760--9779.

\bibitem{liu2024lost}
N.~F. Liu, K.~Lin, J.~Hewitt, A.~Paranjape, M.~Bevilacqua, F.~Petroni, and P.~Liang, ``Lost in the middle: How language models use long contexts,'' \emph{Transactions of the association for computational linguistics}, vol.~12, pp. 157--173, 2024.

\bibitem{peysakhovich2023attention}
A.~Peysakhovich and A.~Lerer, ``Attention sorting combats recency bias in long context language models,'' \emph{arXiv preprint arXiv:2310.01427}, 2023.

\bibitem{maynez2020faithfulness}
J.~Maynez, S.~Narayan, B.~Bohnet, and R.~McDonald, ``On faithfulness and factuality in abstractive summarization,'' \emph{arXiv preprint arXiv:2005.00661}, 2020.

\bibitem{chang2023booookscore}
Y.~Chang, K.~Lo, T.~Goyal, and M.~Iyyer, ``Booookscore: A systematic exploration of book-length summarization in the era of llms,'' \emph{arXiv preprint arXiv:2310.00785}, 2023.

\bibitem{kahneman2011thinking}
D.~Kahneman, \emph{Thinking, fast and slow}.\hskip 1em plus 0.5em minus 0.4em\relax macmillan, 2011.

\bibitem{pan2025survey}
Q.~Pan, W.~Ji, Y.~Ding, J.~Li, S.~Chen, J.~Wang, J.~Zhou, Q.~Chen, M.~Zhang, Y.~Wu, \emph{et~al.}, ``A survey of slow thinking-based reasoning llms using reinforced learning and inference-time scaling law,'' \emph{arXiv preprint arXiv:2505.02665}, 2025.

\bibitem{evans2008dual}
J.~S.~B. Evans, ``Dual-processing accounts of reasoning, judgment, and social cognition,'' \emph{Annu. Rev. Psychol.}, vol.~59, no.~1, pp. 255--278, 2008.

\bibitem{dalgleish2000handbook}
T.~Dalgleish and M.~Power, \emph{Handbook of cognition and emotion}.\hskip 1em plus 0.5em minus 0.4em\relax John Wiley \& Sons, 2000.

\bibitem{goudbeek2010beyond}
M.~Goudbeek and K.~Scherer, ``Beyond arousal: Valence and potency/control cues in the vocal expression of emotion,'' \emph{The Journal of the Acoustical Society of America}, vol. 128, no.~3, pp. 1322--1336, 2010.

\bibitem{wang2023self}
Y.~Wang, Y.~Kordi, S.~Mishra, A.~Liu, N.~A. Smith, D.~Khashabi, and H.~Hajishirzi, ``Self-instruct: Aligning language models with self-generated instructions,'' in \emph{Proceedings of the 61st annual meeting of the association for computational linguistics (volume 1: long papers)}, 2023, pp. 13\,484--13\,508.

\bibitem{bartneck2009measurement}
C.~Bartneck, D.~Kuli{\'c}, E.~Croft, and S.~Zoghbi, ``Measurement instruments for the anthropomorphism, animacy, likeability, perceived intelligence, and perceived safety of robots,'' \emph{International journal of social robotics}, vol.~1, no.~1, pp. 71--81, 2009.

\bibitem{laugwitz2008construction}
B.~Laugwitz, T.~Held, and M.~Schrepp, ``Construction and evaluation of a user experience questionnaire,'' in \emph{Symposium of the Austrian HCI and usability engineering group}.\hskip 1em plus 0.5em minus 0.4em\relax Springer, 2008, pp. 63--76.

\end{thebibliography}

\end{document}